%% file: main.tex
\documentclass[letterpaper]{article} 
\usepackage[submission]{aaai25}  
\usepackage{times}  
\usepackage{helvet}  
\usepackage{courier}  
\usepackage[hyphens]{url}  
\usepackage[pdftex]{graphicx}
\usepackage{graphicx} 
\usepackage{natbib}  
\usepackage{caption} 
\usepackage{algorithm}
\usepackage{algorithmic}

\usepackage{graphicx}
\usepackage[table]{xcolor}

\usepackage{newfloat}
\usepackage{listings}
\DeclareCaptionStyle{ruled}{labelfont=normalfont,labelsep=colon,strut=off} 
\floatstyle{ruled}
\newfloat{listing}{tb}{lst}{}
\floatname{listing}{Listing}
\usepackage{amssymb}
\usepackage{amsmath}
\usepackage{amssymb}
\usepackage{mathtools}
\usepackage{amsthm}
\DeclareMathOperator*{\argmax}{argmax}

\usepackage{caption}

\usepackage{cancel}
\usepackage{graphicx}
\usepackage{color}
\usepackage{booktabs}   
\usepackage{multirow}
\usepackage{colortbl}
\usepackage[table]{xcolor}
\usepackage{graphicx}

\title{Deep Reinforcement Learning for Efficient and Fair Allocation of Healthcare Resources}
\author{
    Written by AAAI Press Staff\textsuperscript{\rm 1}\thanks{With help from the AAAI Publications Committee.}\\
    AAAI Style Contributions by Pater Patel Schneider,
    Sunil Issar,\\
    J. Scott Penberthy,
    George Ferguson,
    Hans Guesgen,
    Francisco Cruz\equalcontrib,
    Marc Pujol-Gonzalez\equalcontrib
}
\affiliations{
    \textsuperscript{\rm 1}Association for the Advancement of Artificial Intelligence\\

    1101 Pennsylvania Ave, NW Suite 300\\
    Washington, DC 20004 USA\\
    proceedings-questions@aaai.org
}

\usepackage{bibentry}

\begin{document}

\maketitle

\begin{abstract}
Scarcity of health care resources could result in the unavoidable consequence of rationing. For example, ventilators are often limited in supply, especially during public health emergencies or in resource-constrained health care settings, such as amid the pandemic. Currently, there is no universally accepted standard for health care resource allocation protocols, resulting in different governments prioritizing patients based on various criteria and heuristic-based protocols. In this study, we investigate the use of reinforcement learning for critical care resource allocation policy optimization to fairly and effectively ration resources. We propose a transformer-based deep Q-network to integrate the disease progression of individual patients and the interaction effects among patients during the critical care resource allocation. We aim to improve both fairness of allocation and overall patient outcomes. Our experiments demonstrate that our method significantly reduces excess deaths and achieves a more equitable distribution under different levels of ventilator shortage, when compared to existing severity- and comorbidity-based methods in use by different governments. Our source code is included in the supplement and will be released upon publication.
\end{abstract}

\input{Sections/intro}

\input{Sections/related_work}
\input{Sections/formulation}

\input{Sections/method}
\input{Sections/exp}

\input{Sections/summary}

\bibliographystyle{aaai25}
\small
\bibliography{aaai25}
\newpage
\onecolumn
\input{Sections/appendix}

\end{document}

%% file: Sections/intro.tex
\begin{figure*}[ht]
\vskip 0.2in
\begin{center}
\includegraphics[width=.85\textwidth]{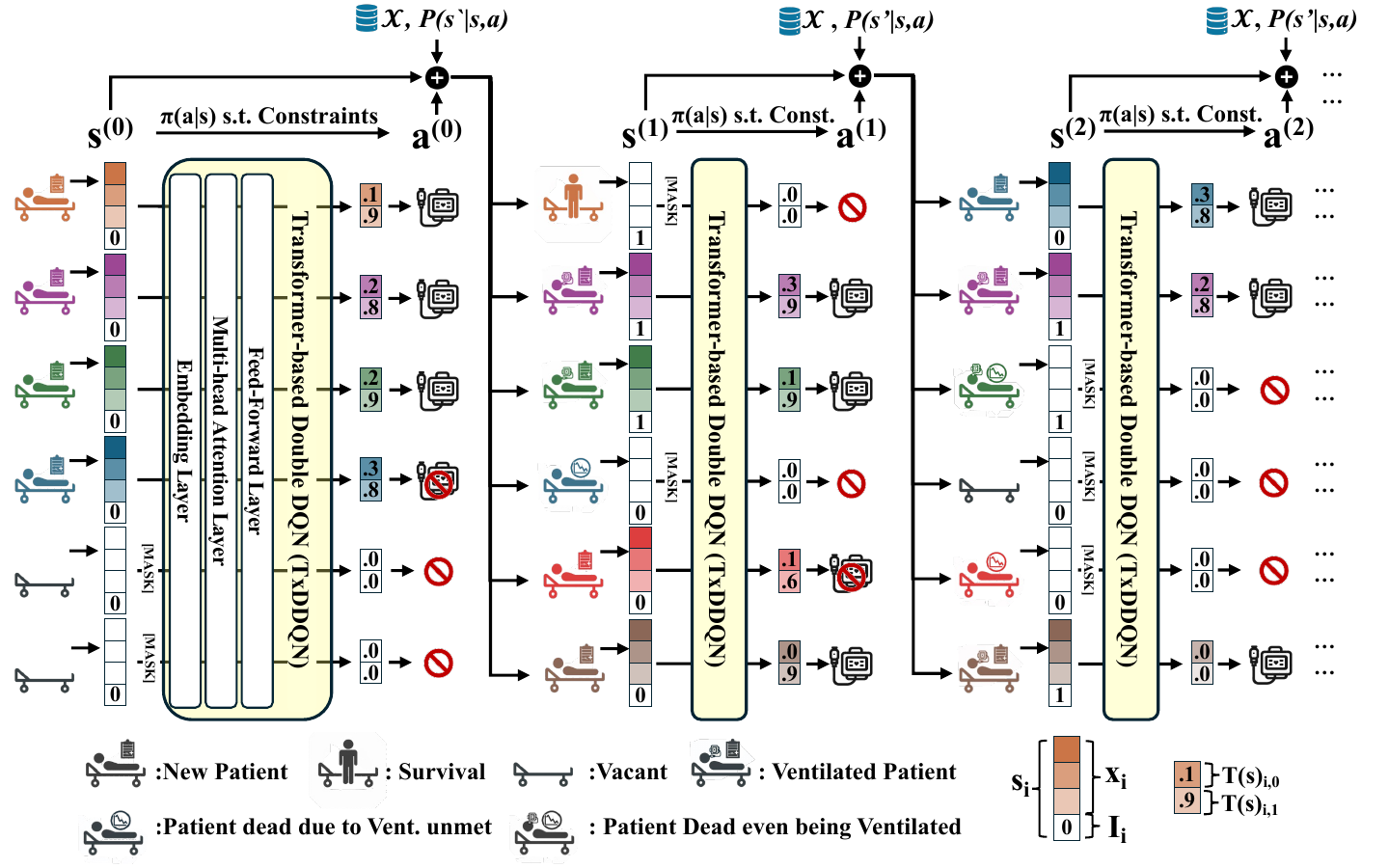}
\caption{An illustration for our study formulation. Each color represents a separate patient.}
\label{fig:study_design}
\end{center}
\vskip -0.2in
\end{figure*}

\section{Introduction}
The Institute of Medicine (IOM) defines Crisis Standards of Care as “a substantial change in usual health care operations and the level of care it is possible to deliver, which is made necessary by a pervasive (e.g., pandemic influenza) or catastrophic (e.g., earthquake, hurricane) disaster” \citep{gostin2012crisis}. These guidelines recognize that pandemics can strain health systems into an absolute scarcity of health care resources and could result in the unavoidable consequence of rationing.

Following the IOM framework, state governments throughout the U.S. have developed allocation protocols for critical care resources during the pandemic \citep{piscitello2020variation}. Consistent with the broad consensus of ethicists and stakeholders \citep{emanuel2020fair}, these protocols aim to triage patients via a pre-specified, transparent, and objective policy. Despite this common general framework, critical details of these protocols vary widely across the U.S. \citep{piscitello2020variation}. The protocols also vary on whether they prioritize younger patients or those without pre-existing medical conditions. For example, the SOFA protocol (e.g., used in NY \citep{VENTILATORALLOCATIONGUIDELINES_2015}) is focused on saving the most lives in the short term and ignores both age and pre-existing medical conditions. In contrast, the multiprinciple protocols (e.g., used in MD and PA) also prioritize younger patients and those without pre-existing medical conditions \citep{biddison2019too}. However, these heuristic-based protocols lack sufficient customization and exhibit considerable variation, underscoring the necessity for evidence-based protocol design using machine learning. The protocols are typically assessed on a daily basis, making it a sequential decision making process and suitable for reinforcement learning (RL).

In this study, we investigate the use of RL for critical care resource allocation policy optimization. Our goal is to learn an optimal allocation policy based on available observations, in order to maximize lives saved while keeping rationing equitable across different races. It is quite natural that sequential allocation decisions on critical care resources can be modeled by a Markov decision process (MDP), and the allocation protocols by RL policies. 

However, application of RL to this problem is nontrivial for practical considerations. In particular, empirical assessment of their performance suggested that most such protocols and their variants used earlier in US states were significantly less likely to allocate ventilators to Black patients \citep{bhavani2021simulation}. In addition, the exact SOFA score is shown to be only modestly accurate in predicting mortality in patients requiring mechanical ventilation \citep{raschke2021discriminant}. There is a vital need to improve both the utility and equity of these allocation protocols to fairly and effectively ration resources to critically ill patients. Our main contributions are summarized as follows:
\begin{itemize}
    \item We are the first to formulate fair health care resource allocation as a multi-objective deep reinforcement learning problem, by integrating the utilitarian and egalitarian objectives into the RL rewards.
    \item We propose a Transformer-based parametrization of the deep Q-network that significantly reduces the complexity of the classical deep-Q network while making allocation decisions based on individual patient disease progression and interactions among patients.
    \item We apply our approach to a large, diverse, multi-hospital real-world clinical datasets. Experiments show that our approach leads to fair allocation of critical care resources among different races, while maintaining the overall utility with respect to patient survival.
\end{itemize}

%% file: Sections/related_work.tex
\section{Related Work}
Health care resource scarcity, especially during pandemics or intensive care, necessitates effective decision-making to optimize outcomes \citep{emanuel2006should,truog2006rationing}. Decision strategies should prioritize optimal use, equity, cost-effectiveness, and crisis management \citep{persad2009principles}, ensuring that they uphold health care as a fundamental human right without discriminating against specific groups. However, implicit or explicit discrimination has permeated healthcare with a long history, presenting numerous instances of biased outcomes \citep{dresser1992wanted,tamayo2003racial,chen2008gender}. 

Due to its inherent nature of comprehending goal-oriented learning and decision-making challenges, RL possesses the potential to generate optimized allocation strategies. However, machine learning approaches, including RL, can be biased towards favoring the accuracies of majority classes at the expense of minority classes, and do not automatically align with fairness objectives \cite{mehrabi2021survey}. In fact, RL may further exacerbate disparities across patient groups by ignorance of fairness considerations \cite{liu2018delayed,ahmad2020fairness,rajkomar2018ensuring,pfohl2021empirical,wang2022comparison,li2022improving}. However, there lack considerations of fairness of RL algorithms in the application of health care resource allocation. In addition, previous research on health resource allocation primarily focuses on either rolling out or one-time distribution of resources, such as vaccines \cite{awasthi2022vacsim,rey2023vaccine, cimpean2023evaluating}, often using multi-armed bandit settings. However, these methods are not suitable for our complex scenario, which requires daily allocation decisions. 

Previous studies used RL to simulate pandemic trajectories, guide lockdown strategies \citep{zong2022reinforcement}, project ventilator needs \citep{bednarski2021collaborative}, and allocate PCR tests for COVID-19 screening \citep{bastani2021efficient}. See survey \cite{yu2021reinforcement} for more comprehensive summary of RL application in health care. However, these methods often overlook long-term fairness, focusing more on efficiency than equity and leaving gaps in addressing disparities in critical resource allocation.

%% file: Sections/formulation.tex
\section{Problem Formulation}
We formulate the ventilator allocation problem as a day-to-day sequential decision problem, as illustrated in Figure \ref{fig:study_design}. \textbf{Our objective is to optimize the triage protocol regarding who should receive health resources under scarcity, with the goal of saving more lives during a health crisis. This optimization by no means influences physicians' medical decisions regarding whether a patient should be ventilated or not.} Therefore, in our formulation and experiments, all patients are prescribed ventilation by physicians but might not receive one due to resource limitation.


\subsection{State Space} 
The state space $\mathcal{S}$ describes the current clinical conditions of all patients in the hospital, as well as their ventilation status. Each state is represented by $s = [x_1,x_2,\cdots, x_N, I_1, I_2, \cdots, I_N]\in\mathcal{S}\subseteq\mathbb{R}^{kN + N}$, where $x\in[0,1]^k\subseteq\mathbb{R}^k$ denotes the current medical condition of the patient on bed $i$, and indicator $I_i\in\{0, 1\}$ denotes whether the bed $i$ has been ventilated. Apart from normal medical conditions of patients, we consider three special conditions: \texttt{Survived}, \texttt{Dead}, and \texttt{Vacant}, corresponding to the cases where the patient in this bed is recovered or dead after ventilation, as well as currently no patient in this bed. They act as terminators for patients or separators between patients in the same bed. Such designs separate different patients on the same bed explicitly, so that the tasks of learning the progression of medical conditions given ventilation and the task of recognizing the end of each patient can be decoupled. The intensive care units have up to $N$ beds. Here we only consider ventilator scarcity, but bed scarcity can be analogously modeled. In the following, we give a rigorous formulation of the RL model for both cases of without and with the consideration of the fairness in distributing ventilators.

With fairness in consideration, we further record the cumulative numbers of total and ventilated patients of different ethnoracial groups in the state vector, denoted by $n_k, m_k\in\mathbb{R}$ respectively, where $k\in\{\text{\textit{B}},\text{\textit{W}},\text{\textit{A}}, \text{\textit{H}}\}$ denotes 4 ethnoracial group in the dataset: non-Hispanic Black, non-Hispanic White, non-Hispanic Asian, Hispanic. Thus, each state $s = [x_1,\cdots, x_N, I_1, \cdots, I_N, n_B, n_W, n_A, n_H, m_B, m_W, m_A,\\ m_H]\in\mathcal{S}\subseteq \mathbb{R}^{kN + N + 8}$ describes the medical and ventilation status of all current patients, as well as the number of cumulative total and ventilated patients of different groups.

\subsection{Action Space}
For both cases, the action space $\mathcal{A}$ is a discrete space denoting whether each bed is on ventilation or not. Let us use 1 to denote ventilate and 0 otherwise, so the action space $\mathcal{A} \subseteq \{0, 1\}^{N}$. Note that we have two constraints on actions: 
\begin{itemize}
    \item Capacity Constraint: $\mathcal{A} \subseteq \{a\in \{0, 1\}^{N} : \sum_{i=1}^N a_i \leq C\}$, where $C$ is the ventilator capacity of the hospital.
    \item Withdrawal Constraint (optional): a patient who has been ventilated must not be withdrawn until they no longer need it or are discharged.: $\mathcal{A} \subseteq \{a\in \{0, 1\}^{N}: a_i = 1 \text{ if } I_i = 1, i= 1, 2, \cdots, N\}$, where $I_i$ in the state information denotes whether current patient on bed $i$ has been ventilated before. We apply this constraint following \cite{bhavani2021simulation}. However, in real clinical practice, the withdrawal of a ventilator from one patient to save another raises ethical issues and has not reached a consensus (e.g., ranging from no mechanism in the Maryland protocol to an explicit SOFA-based approach in the New York protocol). Therefore, we also include the results without this constraint in Appendix A.5 to provide a comprehensive picture across the spectrum, considering both extremes where withdrawal is either considered or not at all.
\end{itemize}
Therefore, the action space is shrunk to 
$\mathcal{A} = \{a \in \{0, 1\}^{N} : \sum_{i=1}^N a_i \leq C \text{ and } \nonumber a_i = 1 \text{ if } I_i = 1, \forall i = 1, 2, \cdots, N\}$.

\subsection{Transition Model}
For ease of notation, we denote the three special conditions  \texttt{Survived}, \texttt{Dead}, and \texttt{Vacant} as $\mathbf{1},\mathbf{-1},\mathbf{0}\in\mathbb{R}^k$ respectively. In the case without fairness consideration, given current state $s = [x_1, x_2, \cdots, x_N, I_1, I_2, \cdots, I_N]$ and action $a = [a_1, a_2, \cdots, a_N]$, it will transit to the next state $s' = [x_1', x_2', \cdots, x_N', I_1', I_2', \cdots, I_N']$ in the following coordinate-wise way: 
\begin{itemize}
    \item If $x_i \neq \mathbf{0}, \mathbf{1}, \mathbf{-1}$, then the patient will transit to \texttt{Dead} condition if not ventilated:
    \begin{align}
    P^{\text{clinical}}_i(x_i'|s,a) = \begin{cases}
        1 & \text{if } a_i = 0, x_i' = \mathbf{-1},\\
        0 & \text{if } a_i = 0, x_i' \neq \mathbf{-1},\\
        p^{\text{on}}(x_i'|x_i) & \text{if } a_i = 1, x_i' \neq \mathbf{0},\\
        0 & \text{if } a_i = 1, x_i' = \mathbf{0}, \nonumber
    \end{cases}
    \end{align}
    where $p^{\text{on}}(x_i' | x_i)$ denotes the probability of a patient transiting from medical condition $x_i$ to $x_i'$ given ventilation. The ventilation status is naively transited $P^{\text{vent}}_i(I_i'|s,a)= 1 \text{ if } I_i' = a_i$ and 0 otherwise. We did not use computational methods to simulate \( p^{\text{on}}(x_i' | x_i) \), as the progression of patients' conditions is high-dimensional and difficult to model. Instead, we only used real clinical trajectories by sampling from real-world clinical databases.

    Following existing clinical literature (e.g., \cite{bhavani2021simulation}), we assume that patients who needed ventilators but did not receive one will die. Thus, there will be no patients \textit{waiting} for a ventilator, as the inability to receive ventilation will result in their immediate deceased and removal from the dataset. However, such an assumption might be over-pessimistic. Therefore, in Appendix A.6, we also explore scenarios where patients not being allocated ventilators does not lead to immediate fatalities, which validates the broader applicability of our proposed methods.

    
    \item If $x_i = \mathbf{0}, \mathbf{1} \text{ or }\mathbf{-1}$, the action does not influence the transition dynamic:
    \begin{align}
    P^\text{clinical}_i(x_i'|s,a) = \begin{cases}
    0 & \text{if } x_i' = \mathbf{1}\text{ or } \mathbf{-1},\\
        1 - q_i(s) & \text{if } x_i' = \mathbf{0},\\
        q_i(s)\cdot \xi(x_i') & \text{if }  x_i' \neq \mathbf{0}, \mathbf{1}, \mathbf{-1}, \nonumber
    \end{cases}
    \end{align}
    where $\xi(\cdot)$ denotes the distribution of the initial medical condition of a patient when admitted to the critical care units, and $q_i(s)$ denotes the probability of a new incoming patient staying in bed $i$. Note $q_i(s)$ depends on how new patients are distributed to empty beds. It satisfies $q_i(s)=0$ if the bed $i$ is already occupied, and $\sum_{i=1}^N q_i(s) = E \sim \text{Poisson}(\Lambda)$ assuming the number of incoming patients $E$ obeys an Poisson distribution with parameter $\Lambda$. 
    The ventilation status does not matter at this point, we can set $P^{\text{vent}}_i(I_i'|s,a)= 1 \text{ if } I_i' = 0$ and 0 otherwise.
\end{itemize}
Given above discussion for transition dynamics of each individual patients/bed, the overall transition probability can be written as 
\begin{align}\label{Eq:transit}
    P = (P^{\text{clinical}}_0, \cdots, P^{\text{clinical}}_N, P^{\text{vent}}_0, \cdots, P^{\text{vent}}_N)
\end{align}

We further consider the progress of the number of cumulative patients and ventilated patients of each ethnic group, whose deterministic transitions are naive by their definitions.

\subsection{Reward}
The reward function consists of the following three parts: 
    \begin{itemize}
        \item Terminal condition $\mathbf{R_t}$: If a patient is discharged alive after being on a ventilator, a positive reward of 1 is given. If a patient requires a ventilator but is not able to receive one, or dies after being on a ventilator, a penalty of -1 is given:
        $R_t(s, a) = \sum_{i=1}^N \mathbf{1}[x_i = \mathbf{1}] - \sum_{i=1}^N \mathbf{1}[x_i = \mathbf{-1}]$.
        \item Ventilation cost $\mathbf{R_v}$: if a ventilator is used, it occurs a small negative reward:$R_v(s, a) = \sum_{i=1}^N a_i$.
        \item Fairness penalty $\mathbf{R_f}$: in the case with fairness consideration, we consider the cumulative total and ventilated patients of different ethnoracial groups. We expect the distribution of ventilators is equitable in terms of the proportion of ventilated patients of all ethnoracial groups. Therefore, a penalty of KL-divergence between the frequency distributions of incoming patients of different ethnoracial groups $\mathcal{D}_n \overset{d}{\sim} [n_B, n_W, n_A, n_H] / (n_B + n_W + n_A + n_H)$ and ventilated patients of different ethnoracial groups $\mathcal{D}_m \overset{d}{\sim}  [m_B, m_W, m_A, m_H] / (m_B + m_W + m_A + m_H)$ is considered: $R_f(s, a) = \hbox{KL}\left(\mathcal{D}_n\large \Vert \mathcal{D}_m\right )$.
    \end{itemize}
Thus, the reward function is given by
\begin{align}\label{eq:reward}
    R(s,a) = R_t(s, a) + \mu \cdot R_v(s, a) + \lambda \cdot R_f(s, a),
\end{align}
where the parameter $\lambda\geq 0$ balances the trade-off between ventilation effectiveness and fairness. In the case without fairness consideration, we set $\lambda = 0$. $\mu$ is a small scalar that controls ventilation costs and can be selected through parameter fine-tuning or clinical expertise.

%% file: Sections/method.tex
\section{Method}
In our formulation, due to the resource constraints imposed through the restricted action sets and the fairness requirements via reward penalties, the interaction effects among the patients need to be considered. However, modeling all patients jointly poses computational and memory challenges due to the combinatorial nature of the action space. Addressing such a formulation necessitates a Q network with dimensions proportional to the size of the state space $|\mathcal{S}|$ and action space $|\mathcal{A}|$, resulting in a complexity up to $O(N \times 2^N)$, where \(N\) represents the ICU bed capacity. This complexity makes it impractical to naively construct and train such a Q network, given the challenges in collecting a sufficient number of data points. In practical terms, the intensive care units of health systems typically have hundreds of beds \((N\geq 100)\), making the computational demands and data requirements for training such a network unattainable. Also, naively concatenating all patients' state vectors introduces an order among patients, which may cause the model to learn artificial factors on that order and potentially bias the model.

To circumvent the computational intractability without losing the consideration of the interaction effects among patients, we therefore propose Transformer-based Q-network parametrization, which inherits the classical Q-learning framework with new parametrization and greedy action selection tailored to our problem structure. In our formulation, the transition of states is partially decomposable because a single patient's clinical conditions depend only on their preceding clinical conditions and ventilation status. Therefore, we can reshape the state from a one-dimensional long vector to a two-dimensional matrix, with the \(i\)-th row as \([x_i, I_i]\) for patient \(i\). In case the fairness features $n_k, m_k$ are also considered, we can replicate them $N$ times and append them to the end of each row. The Q-network is parametrized as $T_\theta: \mathbb{R}^{\text{dim}(\mathcal{S})} \to \mathbb{R}^{N \times 2}$, whose input is the reshaped state matrix. The $i$-th row of the output $T_\theta(s)_{i} \in \mathbb{R}^2$ corresponds to the Q-value contributed by the $i$-th patient given the current state $s$ and the action $a_i$ on the $i$-th patient. We adopt an additive form of the joint Q-value, which estimates the trade-off between effectiveness and fairness when allocating under the constraints and considering the clinical conditions of all patients: $Q(s,a) = \sum_{i=1}^N T_\theta(s)_{i, a_i}$. We leverage the Transformer architecture ~\citep{vaswani2017attention} for $T_\theta$. Each row in the reshaped $s$ are considered as an input patient token. As our input to the model is essentially an orderless set of patients' conditions with indefinite size, Transformers are a natural fit for our problem because Transformers without positional encodings are permutationally-invariant and can handle inputs with arbitrary sizes. Each Transformer layer is composed of a feed-forward layer and a multi-head attention layer. The feed-forward layer acts independently on each element of the input as a powerful feature extractor, while the attention layer is able to capture the interaction effects between all the elements. This quadratic complexity of Transformer architecture also results in a significant reduction in the network's complexity, decreasing it from $O(N \times 2^N)$ to a more manageable $O(N^2)$. We also provide robust empirical results to demonstrate the effectiveness and advantages of our proposed transformer-based parametrization compared to the classical deep Q-network in Appendix A.4.

In this parametrization, the greedy action shall be searched within the valid action space $\mathcal{A}$ to accommodate the withdrawal and capacity constraints: $\pi_\theta(a|s): a^* = \argmax_{a \in\mathcal{A}} \sum_{i=1}^N T_\theta(s)_{i, a_i}$. Under our parametrization and the additive form of the Q-value, this constrained greedy search can be efficiently solved by first allocating ventilators to those who have already been allocated (withdrawal constraint), where $a_i^*=1 \Leftrightarrow I_i=1$. Subsequently, the remaining ($C - \sum_{i=1}^N I_i$) ventilators are allocated to the newly admitted patients for whom the ventilation improvement $d_i=T_\theta(s)_{i, 1} - T_\theta(s)_{i, 0}$ ranks at top-$(C - \sum_{i=1}^N I_i)$ among all their competitors (capacity constraints).

We also develop a simulator, $\text{Simu}( \mathcal{X}; C, \Lambda)$, constructed from real-world clinical trajectories to create the replay buffer for training the proposed model. This simulator maintains the clinical trajectory of each patient while randomizing their relative admission order. $\mathcal{X}$ is a set of clinical trajectories from all patients in the training cohort, with each trajectory being of various lengths covering all the clinical conditions of a patient in the critical care unit. At each time step in the simulator, we sample the initial conditions of $E$ patients from $\mathcal{X}$ without replacement. $E$ is determined by a Poisson distribution with parameter $\Lambda$, inferred from the distribution of ventilator requests in the training cohort. We apply a protocol $\Pi$ to allocate the available $C$ ventilators when the sum of existing patients and newly admitted $E$ patients exceeds the capacity $C$. Patients who are not allocated ventilators or those who are discharged (either alive or deceased) are removed from the ongoing simulator and returned to the sampling pool. Patients who are allocated ventilators progress to the next time step of their own trajectory based on the transition function $p^{on}(x_i'|x_i)$. With this simulator setup, we can generate MDP tuples of indefinite length.

We combine the \textbf{Trans}former q-network parametrization with \textbf{D}ouble-\textbf{DQN} (\textbf{TxDDQN}) ~\citep{van2016deep}, a variant to the original DQN \citep{mnih2013playing} capable of to reduce overestimation. We summarize our proposed model and simulator in Algorithm ~\ref{alg:algo}.

\input{Sections/algorithm}

%% file: Sections/algorithm.tex
\begin{algorithm}[tb]
    \caption{Transformer-based Double Deep Q Network}
    \label{alg:algo}
    \textbf{Input:} $\text{Simu}(\mathcal{X}; C, \Lambda) \text{, discount factor } \gamma \text{, learning rate }  \alpha, \\
    \text{batch size } |\mathcal{B}| \text{, loss function } \mathcal{L}_c \text{,} \text{network update frequency } h, \\
    \text{network update parameter } \tau$\\
    \textbf{Initialize: } Primary network $T_{\theta_0}$; Target network $T_{\theta'}, \theta' \leftarrow \theta_0$; Allocation protocol $\Pi \leftarrow \pi_{\theta_0}$; $\mathcal{D} \leftarrow \Phi$; $s^{(0)}$ by sampling initial medical conditions of $\text{Pois}(\Lambda)$ patients from $\mathcal{X}$\\
    \textbf{for} $e=0 \textbf{ to } E\hfill$ \textit{$\triangleright$ Training loop in epochs}\\
    \hspace*{1em} \textbf{for} $t=0 \textbf{ to } T-1$ \hfill \textit{$\triangleright$ Construct a ring replay buffer}\\
    \hspace*{2em} $a \sim \pi_{\theta_e}(s)$ \\
    \hspace*{2em} \hfill $\triangleright$ $\pi_\theta(a|s): a^* = \argmax_{a \in\mathcal{A}} \sum_{i=1}^N T_\theta(s)_{i, a_i}$\\
    \hspace*{2em} $s' \sim P(s,a)$ \hfill \textit{$\triangleright$ Eq.~\eqref{Eq:transit} for $P(s,a)$}\\
    \hspace*{2em} $\mathcal{D} \leftarrow \mathcal{D} \cup \{(s,a,s',R(s,a)\}$ \hfill \textit{$\triangleright$ Eq.~\eqref{eq:reward} for $R(s,a)$}\\
    \hspace*{1em} $\theta_{e,0} \leftarrow \theta_e$ \\
    \hspace*{1em} \textbf{for} $g=0 \textbf{ to } G-1$ \hfill \textit{$\triangleright$ Gradient steps}\\
    \hspace*{2em} Sample mini-batch $\mathcal{B} \subset \mathcal{D}$ \hfill $\triangleright$ $\mathcal{B} = \{(s, a, s', r\}^{|\mathcal{B}|}$\\
    \hspace*{2em} \textbf{for} $i=0 \textbf{ to } N$ \hfill \textit{$\triangleright$ Search next action $a'^*$}\\
    \hspace*{3em} $d_i = (T_{\theta_{e, g}}(s')_{i, 1} - T_{\theta_{e, g}}(s')_{i, 0})$ \hfill \textit{$\triangleright$ Vent. improv.}\\
    \hspace*{3em} \textbf{if} $I_i'= 1$ \textbf{then} \\
    \hspace*{6em} $a_i' \gets 1$ \hfill \textit{$\triangleright$ Withdrawal const.} \\
    \hspace*{3em} \textbf{elif} $d_i \text{ ranks top- }(C - |I'|) \text{ in } \{d_i \mid I_i'=0\} \textbf{then}$ \\
    \hspace*{6em} $a_i' \gets 1$ \hfill \textit{$\triangleright$ Capacity const.} \\
    \hspace*{3em} \textbf{else } \quad $ a_i' \gets 0$ \\
    \hspace*{2em} $\theta_{e,g+1} \leftarrow \theta_{e,g} - \alpha \cdot \nabla_\theta \mathcal{L}_c [
                \sum_{i=1}^N T_{\theta_{e, g}}(s)_{i,a_i}\textbf{ , } R(s, a)+\\
                \hspace*{6em} \textbf{ }\gamma \cdot \sum_{i=1}^N T_{\theta'}(s')_{i, a_i'}]$ \hfill \textit{$\triangleright$ Policy update}\\
    \hspace*{2em} \textbf{if} $g$ \text{mod} $h = 0$ \textbf{then} $\theta' \leftarrow \tau \cdot \theta_{e,g+1} + (1 - \tau) \cdot \theta'$
    \hspace*{2em} $ \theta_e \leftarrow \theta_{e, G}$
\end{algorithm}

%% file: Sections/exp.tex
\section{Experiments}

\input{tables/table1}

\subsection{Dataset}
The dataset is sourced from a comprehensive and integrated repository of all clinical and research data sources, including electronic health records, pathology data from multiple real-world hospitals and research laboratories.
We collected 11,773 ICU admissions that have been allocated mechanical ventilators between March 15, 2020 and January 15, 2023. We filtered the patients with age between 18 and 95 for consideration. We removed admissions with a ventilator allocation duration exceeding 30 days to eliminate anomalies in the data. We extracted 38 features for each admission, including SOFA components, vital signs, demographics, comorbidities (see Appendix A.1 for the complete list of features and their statistics).  We will share the de-identified dataset upon publication.

We split our data into 3 splits, 5,455 admissions between March 15, 2020 and July 14, 2021 were used as training data; 1,047 admissions between July 15, 2021 and October 14 2021 were used as validation data on which we selected the best hyper-parameters for testing; 5,271 admissions between October 15, 2021 and January 15, 2023 were used as test data. The patient distribution of different races and the ventilator demands on each day are shown in Appendix A.1. We assume that the original dataset obtained from health systems reflects a scenario where there was an abundant supply of ventilators available. We aim to investigate the effectiveness of various protocols in mitigating excess deaths in the event of ventilator shortage.

This setup differed from conventional RL settings, which are trained and evaluated solely on a simulator. We had separate validation and testing sets. This design allowed us to compare the proposed method with existing protocols in real-world hospital operational settings, enhancing the effectiveness and generalizability of our model. Additionally, ventilator demands may fluctuate due to seasonality and outbreaks. A test set spanning a whole year can assess our method's vulnerability to these seasonal request surges.

\subsection{Baseline Protocols}
Our proposed method was compared with the following existing triage protocols:
\textbf{Lottery}: Ventilators are randomly assigned to patients who are in need.
\textbf{Youngest First}: The highest priority is given to the youngest patients.
\textbf{SOFA}: Patients' prioritization is discretized into three levels (0-7: high, 8-11: medium, and 11+: low) with the lottery serving as the tiebreaker.
\textbf{Multiprinciple (MP)}: Each patient is assigned a priority point based on their SOFA score (0-8: 1, 9-11: 2, 12-14: 3, 14+: 4). Patients with severe comorbidities receive an additional 3 points. In case of ties, priority is given to patients in a younger age group (Age groups: 0–49, 50–69, 70–84, and 85+). If ties still exist, a lottery is conducted to determine the final allocation. 
\textbf{Decision Tree (DT)}: Grand-Clément et al. \cite{DBLP:journals/corr/abs-2110-10994} introduced a data-driven decision-tree-based approach for optimizing ventilator allocation. Their protocol factors in the BMI, age, and SOFA score, classifying patients into two priority levels. Admission time is the tie-breaker within the same priority level. 

\subsection{Off-policy and offline training}
The maximum daily demand for ventilators in the validation and testing sets is 85. Therefore, we trained 85 capacity-specific protocol models $C = 1, 2, \cdots, 85$. Ventilator capacity is normalized from [0, 85] to [0\%, 100\%]. We used $\Lambda = 12$ because the number of daily newly admitted patients to the critical care units in our dataset follows a Poisson distribution with $\Lambda = 12$.  Our simulator supports both off-policy and offline RL training settings \cite{levine2020offline}. In the \underline{off-policy} RL setting, we iteratively update $\Pi$ with $\pi_{\theta}$ when constructing the replay buffer. We refer to the off-policy trained model without and with fairness consideration as \textbf{TxDDQN} and \textbf{TxDDQN-fair}. In the \underline{offline} setting, we use existing heuristic-based protocol, MP, as $\Pi$ to create a large training set and do not update the training data during the training process. This offline training process mimics early-stage health crisis operations where resources are allocated using existing protocols, and improvements are sought thereafter. We refer the offline trained model with fairness consideration as \textbf{TxDDQN-fair-off}. Our experiments were conducted on firewall-protected servers. The training time for each capacity-specific model was less than 30 minutes using a single GPU. For hyper-parameter selections and the survival-fairness Pareto frontier, please see Appendix A.2.

\subsection{Evaluation}
We evaluated the performance of our proposed protocol based on normalized survival rates, with the survival number at no shortage in ventilators as 100\% and no patients alive when no ventilator is available as 0\%. Fairness was quantified by comparing the allocation rates across four ethnoracial groups. The allocation rate was calculated by dividing the total number of ventilators allocated by the sum of the ventilators requested. The demographic parity ratio (DPR) \citep{bird2020fairlearn} served as the group metric of fairness, and it is defined as the ratio between the smallest and largest group-level allocation rate. A DPR value close to 1 signifies an equitable allocation, indicating a non-discriminatory distribution among the different groups. Additionally, to gain insights into the impact of protocols under varying shortage levels, we visualized the survival-capacity curve (SCC) and allocation-capacity curve (ACC).
\input{Sections/figures}

\subsection{Results}
Our findings regarding the survival rates for different triage protocols under various levels of ventilator shortage are illustrated in Figure \ref{fig:SCC}. Across all triage protocols, higher ventilator capacities are associated with increased allocation rates, thereby resulting in saving more lives. The Area Under the Survival-Capacity Curve (AUSCC) serves as an indicator of the overall performance in terms of life-saving abilities under different levels of ventilator shortage. Similarly, the Area Under the Allocation-Capacity Curve (AUACC) reflects the overall performance of ventilator utilization rates under varying levels of shortage. In the left Panel of Figure \ref{fig:SCC}, our proposed TxDDQN models exhibit higher AUSCC compared to all other baseline protocols, demonstrating the superiority of our models in terms of life-saving efficacy. Likewise, in the right Panel, TxDDQN models also demonstrate higher AUACC compared to other baselines, indicating their superiority in terms of ventilator utilization.

Figure \ref{fig:ACC} showcases the allocation-capacity curves for different ethnoracial groups and triage protocols in Panels A through H. Only Lottery, TxDDQN-fair and TxDDQN-fair-off model exhibit minimal disparities, but TxDDQN-fair and TxDDQN-fair-off surpasses Lottery in terms of allocation rates and life-saving efficacy. Conversely, all other triage protocols display a preference for specific ethnoracial groups. For example, Youngest and MP favor Hispanic, while SOFA and TxDDQN favor White.

We conducted a detailed analysis in Table 1 under the scenario where approximately 50\% of ventilators are unavailable. Our results show that TxDDQN-fair achieves the highest survival rate and allocation rate among the triage protocols. It ranks second in terms of DPR, following closely behind the Lottery protocol. These findings confirm the effectiveness of our TxDDQN-fair approach in improving both survival rates and fairness simultaneously. Importantly, the inclusion of fairness rewards does not compromise its life-saving capabilities compared to TxDDQN. We also did not observe differences between TxDDQN-fair and TxDDQN-fair-off in their life-saving abilities and fairness. This confirms that our proposed methods are adaptable to both offline and off-policy training, and provide the foundation for safe deployment in the early stages of health crises. Our proposed model also demonstrates enhanced fairness and life-saving outcomes in two additional settings: i) where withdrawal constraint is removed (Appendix A.5), and ii) the shortage of ventilators does not lead to immediate deaths (Appendix A.6). We provide ablation studies in Appendix A.3.

%% file: tables/table1.tex
\begin{table*}[!ht]
\footnotesize 
\setlength{\tabcolsep}{1mm}
\centering
\label{tab:tab2}
\begin{tabular}{@{}l|l|l|lllll@{}}
\toprule
                  & \textbf{Survival} & \textbf{Fairness} & \multicolumn{5}{l}{\textbf{Allocation Rates}}                            \\ \cmidrule(l){2-8} 
                  & Survival, \%            & DPR, \%           & Overall, \%  & Asian, \%    & Black, \%    & Hispanic, \% & White, \%    \\ \midrule
\textbf{Lottery}    & 75.17 \textpm \scriptsize{0.45} & \textbf{96.89 \textpm \scriptsize{1.55}} & 75.60 \textpm \scriptsize{0.28} & 74.95 \textpm \scriptsize{1.81} & 75.65 \textpm \scriptsize{1.10} & 76.08 \textpm \scriptsize{0.90} & 75.68 \textpm \scriptsize{0.33} \\
\textbf{Youngest} & 77.24 \textpm \scriptsize{0.07} & 86.39 \textpm \scriptsize{0.27} & 75.65 \textpm \scriptsize{0.09} & 75.59 \textpm \scriptsize{0.50} & 81.73 \textpm \scriptsize{0.31} & 84.44 \textpm \scriptsize{0.30} & 72.94 \textpm \scriptsize{0.12} \\
\textbf{SOFA}     & 80.88 \textpm \scriptsize{0.32} & 92.37 \textpm \scriptsize{1.01} & 77.92 \textpm \scriptsize{0.25} & 75.06 \textpm \scriptsize{1.44} & 73.58 \textpm \scriptsize{0.77} & 78.66 \textpm \scriptsize{1.24} & 79.10 \textpm \scriptsize{0.45} \\
\textbf{DT}       & 76.16 \textpm \scriptsize{0.01} & 94.35 \textpm \scriptsize{0.01} & 75.81 \textpm \scriptsize{0.01} & 79.59 \textpm \scriptsize{0.00} & 75.14 \textpm \scriptsize{0.01} & 77.34 \textpm \scriptsize{0.00} & 75.58 \textpm \scriptsize{0.00} \\ 
\textbf{MP}       & 81.99 \textpm \scriptsize{0.14} & 90.68 \textpm \scriptsize{0.95} & 78.34 \textpm \scriptsize{0.20} & 77.91 \textpm \scriptsize{1.48} & 74.33 \textpm \scriptsize{0.66} & 81.96 \textpm \scriptsize{0.56} & 78.81 \textpm \scriptsize{0.34} \\ \midrule
\textbf{TxDDQN}    & 84.76 \textpm \scriptsize{0.24} & 86.91 \textpm \scriptsize{3.45} & \underline{81.80 \textpm \scriptsize{0.26}} & 78.03 \textpm \scriptsize{0.99} & 72.48 \textpm \scriptsize{0.74} & 81.71 \textpm \scriptsize{0.44} & 84.45 \textpm \scriptsize{0.22} \\
\textbf{TxDDQN-fair-off} & \underline{85.29 \textpm \scriptsize{0.18}} & \underline{95.27 \textpm \scriptsize{1.42}} & \textbf{81.96 \textpm \scriptsize{0.15}} & 80.50 \textpm \scriptsize{2.23} & 80.07 \textpm \scriptsize{1.98} & 81.44 \textpm \scriptsize{1.05} & 82.70 \textpm \scriptsize{0.34} \\
\textbf{TxDDQN-fair} & \textbf{85.41 \textpm \scriptsize{0.23}} & \underline{95.24 \textpm \scriptsize{1.65}} & \underline{81.90 \textpm \scriptsize{0.24}} & 80.01 \textpm \scriptsize{1.79} & 79.95 \textpm \scriptsize{1.05} & 81.26 \textpm \scriptsize{1.24} & 82.80 \textpm \scriptsize{0.42} \\ \bottomrule
\end{tabular}%
\caption{Impact of triage protocols on survival, fairness, and allocation rates with limited ventilators (B = 40) corresponding to about 50\% scarcity. Fairness metric is demographic parity ratio (DPR, the ratio between the smallest and the largest allocation rate across patient groups,  100\% indicating non-discriminative). Standard deviations are from 10 experiments with different seeds. We bold the protocols with the highest survival rate, DPR, and overall allocation rates. We underline the protocols that fall within one standard deviation of the best result.}
\end{table*}

%% file: Sections/figures.tex
\begin{figure*}[!t]
    \centering
    \includegraphics[width=\textwidth]{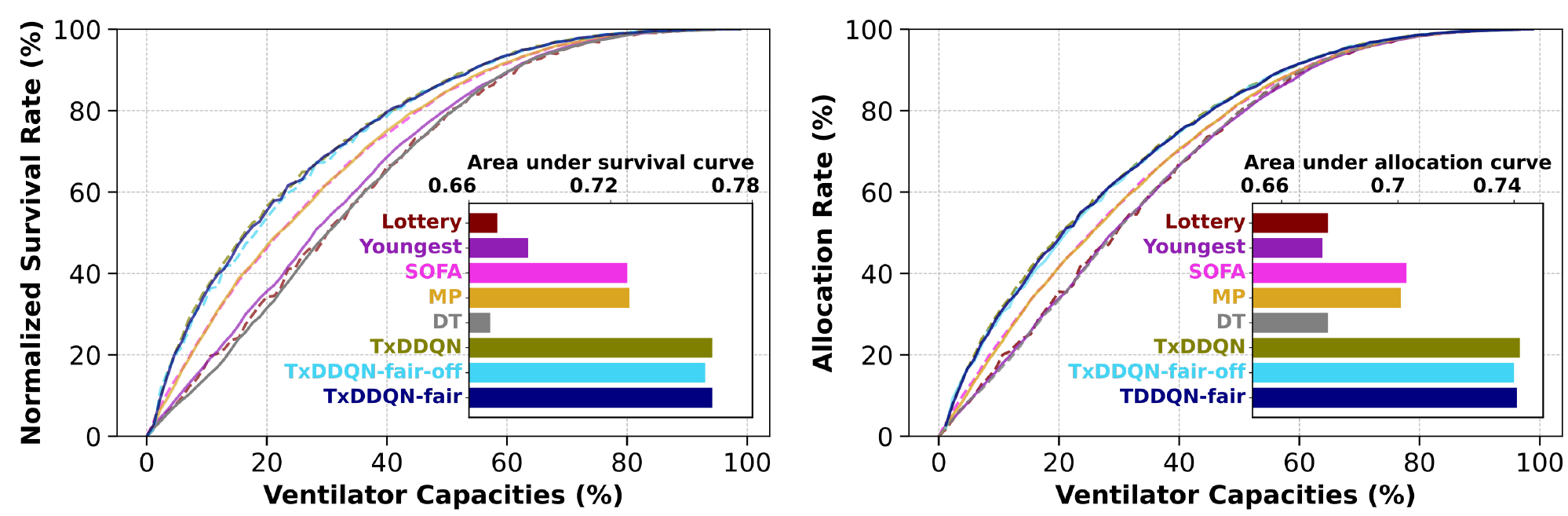}
    \caption{Impact of triage protocols on survival rates and allocation rates under varying levels of ventilator shortages. The maximum daily demand for ventilators in the testing set is considered as full capacity (100\%). We scale the number of survivors to a range of [0, 100\%] to represent the survival rate. The allocation rate is calculated by dividing the total number of ventilators allocated by the total number of the ventilators requested. The bar plot associated with each panel indicates the area under the survival-capacity curve and allocation-capactiy curve, respectively, where a larger value indicates that the protocol can save more lives across different levels of shortages. Notably, the MP and SOFA curves exhibit overlap, indicating similar allocation patterns. Similarly, the lottery and youngest curves show close proximity, as do our three TxDDQN configurations.}
    \label{fig:SCC}
\end{figure*}

\begin{figure*}[!t]
    \centering
    \includegraphics[width=\textwidth]{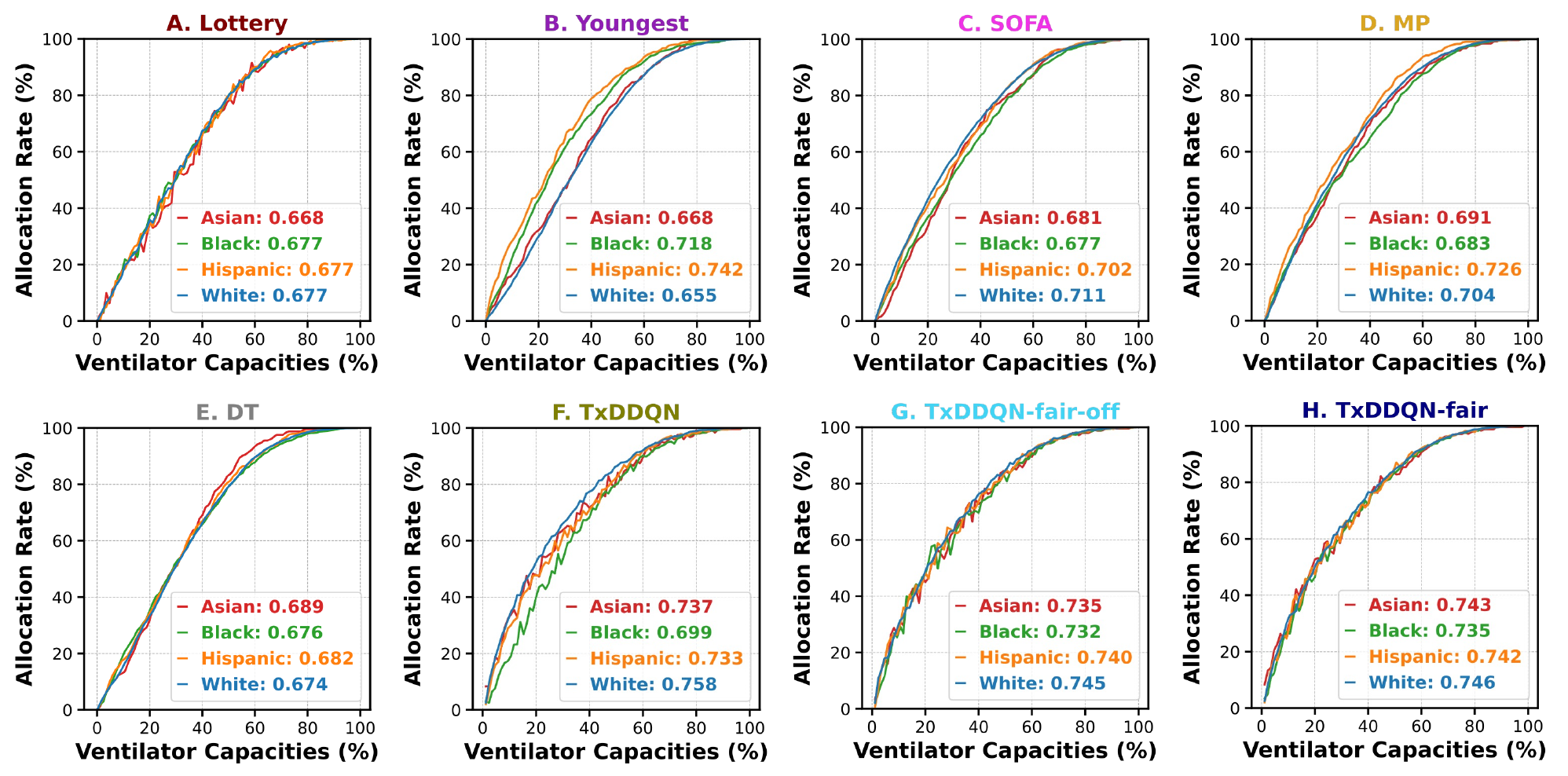}
    \caption{Allocation rates across protocols and ethnoracialgGroups. Each panel illustrates how allocation rates vary by ethnoracial group under different protocols. The numbers in the legend indicate the area under the allocation-capacity curve (AUACC).}
    \label{fig:ACC}
\end{figure*}

%% file: Sections/summary.tex
\section{Summary}
In this study, we formulated fair health care resource allocation as a multi-objective deep reinforcement learning problem. 
Our proposed model outperforms existing protocols used by different states in the U.S., by saving more lives and achieving a more equitable allocation of health resources, as demonstrated by experiments on real-world data.
We refer the readers to Appendix A.7 for a detailed discussion on the limitations and future directions of this study.

%% file: Sections/appendix.tex
\setcounter{table}{0}
\setcounter{figure}{0}
\renewcommand{\thetable}{A\arabic{table}}
\renewcommand{\thefigure}{A\arabic{figure}}

\section{Appendix}
We organize the appendix as follows. Section A.1 provides descriptive statistics of the dataset used in this study. In Section A.2, we provide additional details on the hyperparameters and model framework used in our experiments, as well as a survival-fairness Pareto frontier. In section A.3, we provide ablation studies and results from more baseline models. We conduct empirical comparisons between classical Q network and transformer-based parametrization in their life-saving abilities in section A.4. Section A.5 presents the results of the setting when daily reassessment for ventilator allocation is taken into account. In Section A.6, we discuss the scenario in which the lack of allocation of ventilators does not lead to immediate fatality. Section A.7 summarizes the limitations of this study and outlines potential future directions. Furthermore, we have created separate files containing the codebase for this study in the supplemental materials, which will also be released on GitHub upon acceptance. We will share the de-identified dataset upon publication of the paper.


\newpage
\subsection{A.1 Dataset}\label{sec:dataset}
\begin{table}[!htbp]
\caption{ Summary descriptive statistics of the variables in our dataset. (N = 11773). Binary variables are presented using positive numbers and percentages, while continuous and ordinal variables are described using means and standard deviations. }
\centering
\begin{tabular}{@{}lllrr@{}}
\toprule
\multicolumn{3}{l}{\textbf{Variables}}                                  & \multicolumn{1}{l}{N / Mean} & \multicolumn{1}{l}{\% / Std} \\ \midrule
\multicolumn{5}{l}{\textbf{Demographics}}                                 \\
   & \multicolumn{4}{l}{Sex}                                              \\
   &                      & Female                   & 4630    & 39.30\%  \\
   &                      & Male                     & 7143    & 60.70\%  \\
   & \multicolumn{4}{l}{Race/Ethnicity}                                   \\
   &                      & Asian                    & 479     & 4.07\%   \\
   &                      & Black                    & 1736    & 14.75\%  \\
   &                      & Hispanic                 & 1274    & 10.82\%  \\
   &                      & White                    & 7509    & 63.78\%  \\
   & \multicolumn{2}{l}{Age (yr)}                    & 62      & 15.49    \\
\multicolumn{5}{l}{\textbf{Vital Signs}}                                  \\
   & \multicolumn{2}{l}{Pulse (BPM)}                 & 90.3    & 22.18    \\
   & \multicolumn{2}{l}{SpO2 (\%)}                   & 96.9    & 5.03     \\
   & \multicolumn{2}{l}{Respirations (BPM)}          & 21.2    & 8.15     \\
   & \multicolumn{2}{l}{BMI (kg/m²)}                 & 29.6    & 8.74     \\
   & \multicolumn{2}{l}{Systolic BP (mmHg)}          & 122.1   & 27.64    \\
   & \multicolumn{2}{l}{Diastolic BP (mmHg)}         & 66.7    & 18.64    \\
   & \multicolumn{2}{l}{Temperature (°F)}            & 97.8    & 1.94     \\
\multicolumn{5}{l}{\textbf{SOFA scores}}                                  \\
   & \multicolumn{2}{l}{Respiratory}                 & 1.9     & 1.17     \\
   & \multicolumn{2}{l}{Coagulation}                 & 0.7     & 0.97     \\
   & \multicolumn{2}{l}{Hepatic}                     & 0.5     & 0.88     \\
   & \multicolumn{2}{l}{Cardiovascular}              & 2.7     & 1.40     \\
   & \multicolumn{2}{l}{Neurological}                & 2.5     & 1.43     \\
   & \multicolumn{2}{l}{Renal}                       & 0.9     & 1.18     \\
\multicolumn{5}{l}{\textbf{Comorbidities}}                                \\
   & \multicolumn{2}{l}{Acute Myocardial Infarction} & 2622    & 22.27\%  \\
   & \multicolumn{2}{l}{Congestive Heart Failure}    & 4716    & 40.06\%  \\
   & \multicolumn{2}{l}{Peripheral Vascular Disease} & 3387    & 28.77\%  \\
   & \multicolumn{2}{l}{Cerebrovascular Disease}     & 3350    & 28.45\%  \\
   & \multicolumn{2}{l}{Dementia}                    & 577     & 4.90\%   \\
            & \multicolumn{2}{l}{Chronic Obstructive Pulmonary Disease} & 3673                         & 31.20\%                      \\
   & \multicolumn{2}{l}{Rheumatic Disease}           & 525     & 4.46\%   \\
   & \multicolumn{2}{l}{Peptic Ulcer Disease}        & 596     & 5.06\%   \\
   & \multicolumn{2}{l}{Mild Liver Disease}          & 1855    & 15.76\%  \\
   & \multicolumn{2}{l}{Diabetes}                    & 4132    & 35.10\%  \\
            & \multicolumn{2}{l}{Diabetes with Wound Complications}     & 2391                         & 20.31\%                      \\
   & \multicolumn{2}{l}{Hemiplegia or Paraplegia}    & 1091    & 9.27\%   \\
   & \multicolumn{2}{l}{Renal Disease}               & 3673    & 31.20\%  \\
   & \multicolumn{2}{l}{Cancer}                      & 2260    & 19.20\%  \\
            & \multicolumn{2}{l}{Moderate or Severe Liver Disease}      & 872                          & 7.41\%                       \\
   & \multicolumn{2}{l}{Metastatic Cancer}           & 1173    & 9.96\%   \\
   & \multicolumn{2}{l}{AIDS}                        & 75      & 0.64\%   \\
   & \multicolumn{2}{l}{COVID-19}                    & 1436    & 12.20\%  \\
\multicolumn{5}{l}{\textbf{Outcome}}                                      \\
   & \multicolumn{2}{l}{Death}                       & 2980    & 25.31\%  \\ \bottomrule
\end{tabular}
\end{table}
\clearpage
\newpage
\input{tables/table3}
\begin{figure*}[h!]
\vskip 0.2in
\begin{center}
\centerline{\includegraphics[width=\columnwidth]{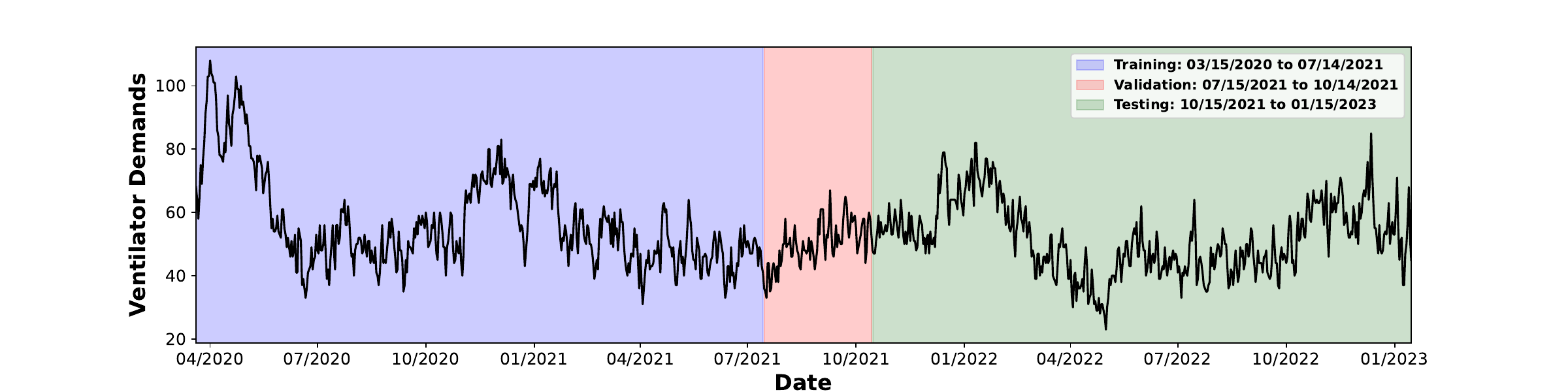}}
\caption{Daily Ventilator Demands}
\label{fig:fig2}
\end{center}
\vskip -0.2in
\end{figure*}
\clearpage
\newpage

\subsection{A.2 Hyper-parameters}\label{sec:hp}
\begin{table}[!ht]
\caption{Hyperparameters for model training details, TxDDQN framwork and the markov decision process in the study formulation}
\centering
\begin{tabular}{@{}lllr@{}}
\toprule
\multicolumn{4}{l}{\textbf{Hyper-parameters}}              \\ \midrule
 & \multicolumn{3}{l}{\textbf{Model training}}             \\
 &  & Batch size $|\mathcal{B}|$                          & 32            \\
 &  & Learning rate $\alpha$                        & 3e-5          \\
 &  & Steps per update $h$                    & 500           \\
 &  & Epochs $E$                    & 60           \\
 &  & Gradient steps $G$                       & 1000         \\ 
 &  & Loss function $\mathcal{L}_c$                    &    \textit{torch.nn.SmoothL1Loss}        \\
 \cmidrule(l){2-4} 
 & \multicolumn{3}{l}{\textbf{TxDDQN}}                      \\
 &  & Hidden size                          & 1024          \\
 &  & Embedding dim                        & 1024          \\
 &  & Attention head                       & 16            \\
 &  & Max seq length                       &  $C+2\Lambda$ \\
\cmidrule(l){2-4} 
 & \multicolumn{3}{l}{\textbf{MDP}}                        \\
 &  & Scalar for ventilator cost ($\mu$)      & -0.1          \\
 &  & Scalar for fairness penalty ($\lambda$) & 1e3          \\
 &  & Discount factor ($\gamma$)                     & 0.95          \\
 &  & Reply buffer size                    & 16000         \\ \bottomrule
\end{tabular}
\end{table}
\newpage
\begin{figure}[hb]
    \centering
    \includegraphics[width=.7\textwidth]{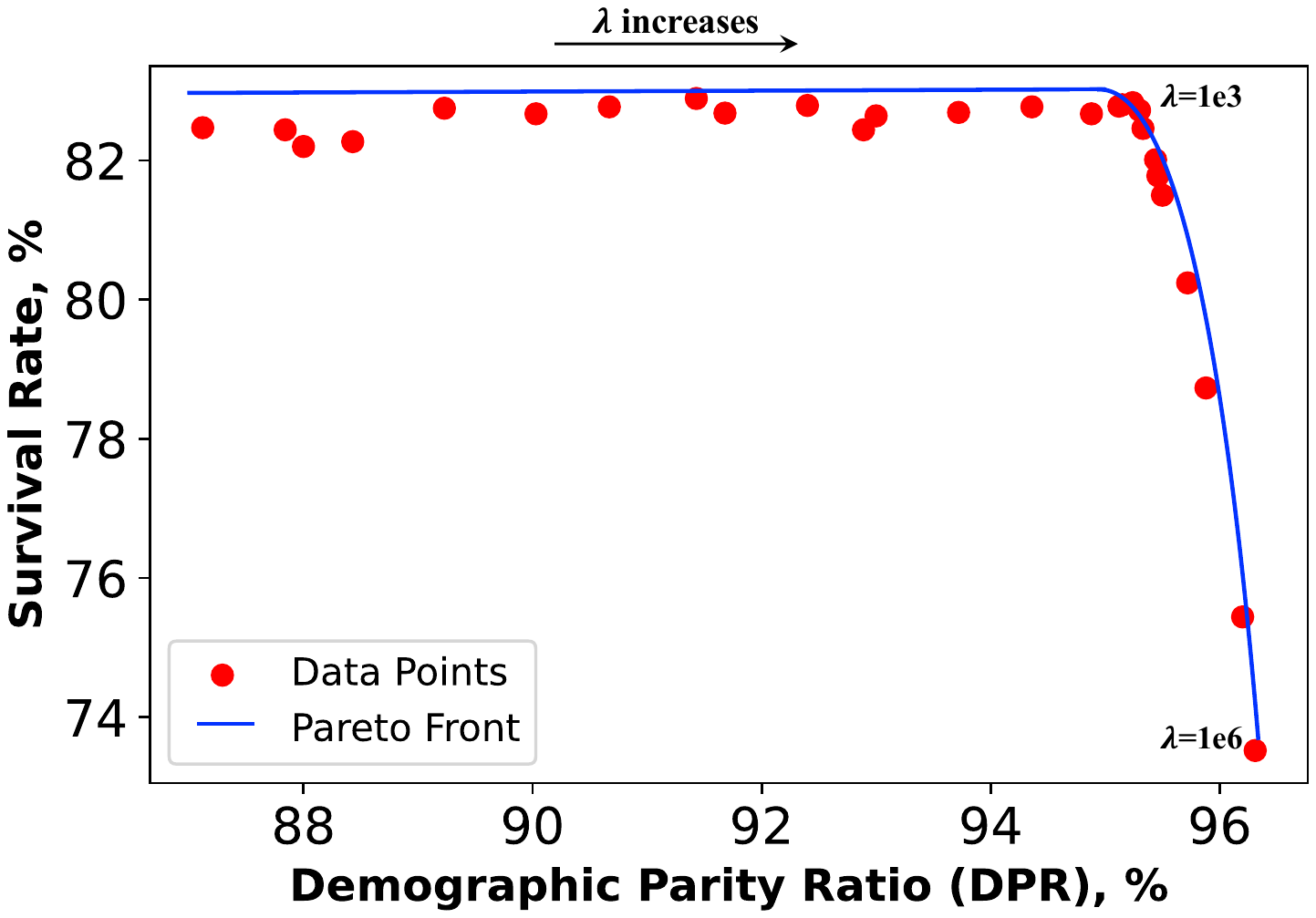}
    \caption{The survival-fairness Pareto frontier was examined under a 50\% shortage of ventilators. Each data point corresponds to the outcomes from distinct values of $\lambda$, which balance the trade-off between allocation effectiveness and fairness. As $\lambda$ increases (red dots move to the right), fairness can be enhanced until a turning point is reached, at which the survival rate begins to decline. If $\lambda$ is increased to an exceedingly large value (e.g., 1e6), the model will tend to favor a protocol that prioritizes fairness without due consideration for life-saving. We reported the results from the turning point ($\lambda$ = 1e3), where the model is enhanced in fairness without compromising the survival rate.}
    \label{fig:pf}
\end{figure}

\newpage

\subsection{A.3 Ablation studies and more baselines}\label{sec:ablation}
In this section, we report additional ablation studies and baselines besides the ablation experiments on rewards in the main text. See Table \ref{tab:aaba} for survival and fairness results with limited ventilators (B = 40) corresponding to about 50\% scarcity.

\textbf{TxDDQN, TxDDQN-\cancel{$\mathbf{R_v}$}, TxDDQN-\cancel{$\mathbf{R_t}$}} Our TxDDQN-fair model considers three reward components. We conduct ablation studies to analyze the impact of each reward component. In these three models, the rewards associated with fairness penalty, ventilation cost, and terminal condition are respectively removed. The results indicate that if the ventilation cost or terminal conditions are ignored, the model can achieve a fair distribution but at the expense of reduced life-saving ability. This finding underscores the superiority of our proposed model in achieving both optimized and equitable allocation.

\textbf{TxDDQN-offline-YF, TxDDQN-offline-SOFA, TxDDQN-offline-MP}
In training the TxDDQN-fair model, we use an off-policy setting by pre-collecting a fixed set of trajectories and storing them in a replay buffer. These trajectories are generated by deploying our training models within a simulator. We iteratively update the training model and collect new trajectories using the updated model. For this offline setting experiment, we collect trajectories using different baseline heuristic protocols. Notably, the data collection occurs only once at the beginning, and the model is subsequently trained on this pre-collected data. This process mimics early-stage health crisis operations where resources are allocated using existing protocols, and improvements are sought thereafter. We report the TxDDQN-offline-MP model in the main text. We report here the model trained on the trajectories using the remaining three heuristics protocols. The results show that models trained with datasets collected using protocols other than the lottery achieve comparable results to TxDDQN-fair. Since the lottery protocol promotes fair allocation, actions leading to inequitable allocations are less explored in the replay buffer and thus not adequately penalized by the model. Consequently, TxDDQN-offline-lottery yields similar survival rates but less equity compared to our TxDDQN-fair model.

\textbf{DDQN-individual} In this experiment, our focus is on modeling the Markov decision process for individual patients using a double DQN network. Unlike the TxDDQN model, which considers the dynamic needs of all patients in the ICU unit at a given time, we specifically examine the trajectory progression of a single patient. Similarly to other baseline approaches, we assign a priority score to each patient based on the expected cumulative rewards when allocating the ventilator to them. However, since our emphasis is on individual patients, the group fairness reward cannot be taken into account in this particular experiment. The result indicates that this method has limited effectiveness in improving the survival rate compared to the baseline protocol. Additionally, it also cannot ensure equitable allocation.

\textbf{TDQN-fair} Our TxDDQN-fair model addresses the issue of overestimation of action values and stabilizes the training processes by leveraging the double deep Q network. However, in this specific experiment, we use a single neural network that simultaneously estimates both the current and target action values. This configuration achieves comparable results to TxDDQN-fair; however, it exhibits larger fluctuations in both survival and fairness, as indicated by a larger standard deviation.

\textbf{TxDDQN-lottery} In the training of the TxDDQN-fair model, we collect a fixed-length set of trajectories in advance and stored them in a replay buffer. These trajectories are obtained by deploying our proposed models for ventilator allocation. We iteratively repeat the process of training the model and collecting new trajectories using the deployed model. For this experiment, we collect the trajectories specifically when using the lottery protocol. It's important to note that the data collection process was executed only once at the beginning, and the model was subsequently trained on this pre-collected experience. Since the lottery protocol inherently promotes a fair allocation, the actions leading to inequitable allocations are not penalized adequately. This experiment yields similar survival rates but less equity compared to our TxDDQN-fair model.

\textbf{TxDDQN-early} This experiment is designed to simulate the scenario that occurred at the initial stage of the pandemic in the real world, where only a limited amount of allocation experience data was available. Unlike in the training of TxDDQN-fair, where we sampled the training data from the patients' trajectories spanning the first 16 months of the pandemic, this experiment focuses on sampling data solely from the first 3 months of the pandemic (03/15/2020 - 06/14/2020). Despite reducing the training set to approximately 20\%, this configuration continues to outperform all baseline protocols in terms of survival rates and promotes an equitable distribution of ventilators.

\begin{table}[!ht]
\centering
\caption{Ablation studies and more baselines}
\begin{tabular}{@{}l|l|l@{}}
\toprule
Model           & Survival, \% & DPR, \%      \\ \midrule
\textbf{TxDDQN-fair}      & \textbf{85.41 ± 0.23} & \textbf{95.24 ± 1.65} \\
TxDDQN & 84.76 ± 0.24 & 86.91 ± 3.45 \\
TxDDQN-\cancel{$\mathbf{R_v}$} & 84.60 ± 0.27 & 94.47 ± 1.84 \\
TxDDQN-\cancel{$\mathbf{R_t}$} & 76.02 ± 0.24 & 95.32 ± 1.26 \\
TxDDQN-offline-lottery   & 85.30 ± 0.29 & 91.73 ± 2.01 \\
TxDDQN-offline-YF   & 85.24 ± 0.17 & 94.70 ± 1.19 \\
TxDDQN-offline-SOFA   & 85.14 ± 0.27 & 92.15 ± 1.92 \\
DDQN-individual & 83.39 ± 0.17 & 88.91 ± 0.84 \\
TDQN-fair       & 84.32 ± 0.53 & 94.54 ± 2.27 \\
TxDDQN-early     & 83.70 ± 0.19 & 94.58 ± 1.24 \\ \bottomrule
\end{tabular}
\label{tab:aaba}
\end{table}

\newpage

\subsection{A.4 Empirical comparisons between classical Q network and transformer-based parametrization}\label{sec:emp}
Our proposed transformer-based parametrization significantly reduces complexity from $O(N\times2^N)$ to a more manageable $O(N^2)$. We aim to investigate whether this simplification compromises the model's life-saving abilities. Establishing an approximation bound for this inherently complex combinatorial optimization problem is highly intriguing yet challenging. Therefore, we rely on empirical results to demonstrate that our transformer-based parametrization, despite its reduced complexity, is comparable to the classical Q network.

These empirical comparisons utilize the simulator $\text{Simu}(\mathcal{X}; C, \Lambda)$ as described in the main text, albeit with smaller values of capacity $C$ and newly admitted patients $\Lambda$ to ensure the classical Q network, which has a complexity of $O(N\times2^N)$, remains manageable. We compare the transformer-based parametrization with the classical Q network under off-policy training conditions across various levels of $N=C+2\Lambda$ for the same number of gradient steps.

The empirical results in Table \ref{tab:emprical} show that our transformer-based parametrization achieves a competitive survival rate compared to the classical Q network when $N$ < 12. However, as $N$ increases, the performance of the classical Q network deteriorates, possibly due to insufficient training data or an excessively large action space. When $N$ reaches 20, the parameters exceed 1 billion and become unmanageable within a single GPU setup.

This empirical experiment demonstrates the effectiveness of our proposed parametrization in reducing complexity without sacrificing the life-saving ability of the models.

\begin{table}[!h]
\centering
\caption{Empirical comparisons between classical Q network and transformer-based parametrization in life-saving ability.}
\label{tab:emprical}
\resizebox{0.66\textwidth}{!}{%
\begin{tabular}{@{}cccccc@{}}
\toprule
\multicolumn{1}{l}{N} &
  \multicolumn{1}{l}{C} &
  \multicolumn{1}{l}{$\Lambda$} &
  \multicolumn{1}{l}{\begin{tabular}[c]{@{}l@{}}\# of Parameters in \\ Classical DQN\end{tabular}} &
  \multicolumn{1}{l}{\begin{tabular}[c]{@{}l@{}}Survival Rate \\ from DQN\end{tabular}} &
  \multicolumn{1}{l}{\begin{tabular}[c]{@{}l@{}}Survival Rate \\ from TxDDQN\end{tabular}} \\ \midrule
6  & 4  & 1 & 1.4 M  & 0.927 & 0.932 \\
8  & 4  & 2 & 1.7 M  & 0.697 & 0.701 \\
10 & 6  & 2 & 2.5 M  & 0.917 & 0.913 \\
12 & 6  & 3 & 5.8 M  & 0.770 & 0.772 \\
14 & 8  & 3 & 18.4 M & 0.879 & 0.913 \\
16 & 8  & 4 & 68.9 M & 0.797 & 0.846 \\
18 & 8  & 5 & 270 M  & 0.775 & 0.793 \\
20 & 10 & 5 & 1076 M & OOM   & 0.869 \\ \bottomrule
\end{tabular}%
}
\end{table}

\newpage
\subsection{A.5 Allocation with daily reassessment}\label{sec:reasses}
In the main text, we discuss the circumstances where patients should not have their ventilators withdrawn once allocated until the end of their ICU course due to the scarcity of resources. However, in real-world settings, there may be situations where the withdrawal of mechanical ventilation from one patient to give it to another becomes necessary during a ventilator shortage, particularly when a patient with higher priority requires the ventilator. \cite{piscitello2020variation} summarized that 22 guidelines in the United States provide instructions on when and what criteria should be used to reassess the priority of a patient. However, a consensus has not been reached on this matter. In the main results, we take a conservative approach by emphasizing that the allocation of ventilation resources from one patient to another should be determined solely by physicians, considering the potential ethical concerns involved. Here, we present the results of our experiment, which allows for daily reassessment, meaning that ventilators are allocated each day based on predefined protocols. Patients who have been already receiving ventilation do not receive priority solely based on their current possession of ventilators. This experiment serves as evidence that our proposed methods also work effectively when reassessment is taken into account, which provide a comprehensive picture across the spectrum connecting both extremes. In real clinical practice, physicians and stakeholders can adjust the frequency and criteria for reassessment based on their expertise and judgment.

\begin{figure}[!t]
    \centering
    \includegraphics[width=\textwidth]{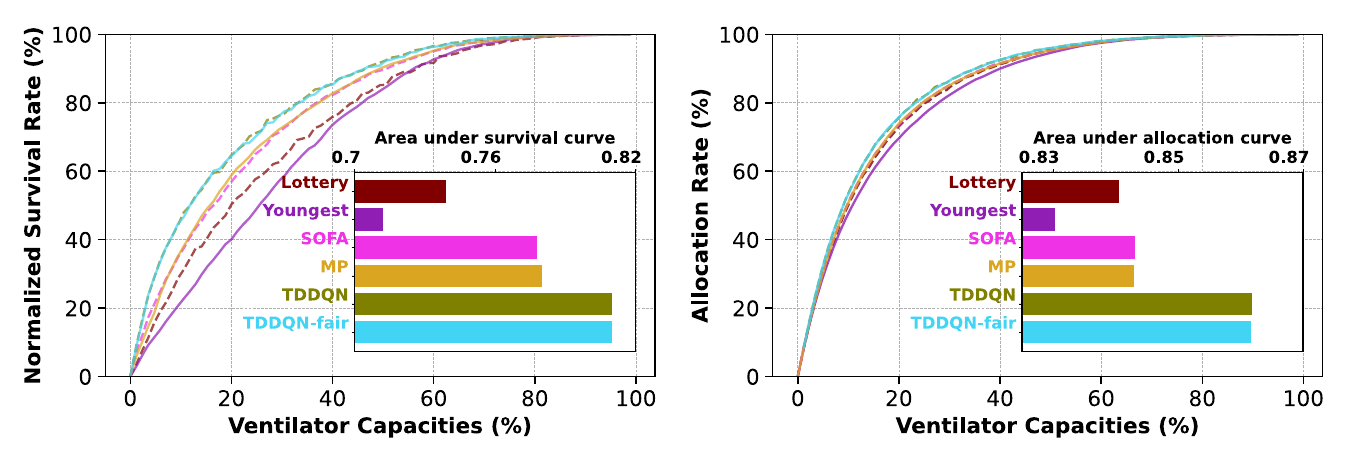}
    \caption{Impact of triage protocols on survival rates and allocation rates under varying levels of ventilator shortages, when daily reassement is considered. The maximum daily demand for ventilators in the testing set is considered as full capacity (100\%). We scale the number of survivors to a range of [0, 100\%] to represent the survival rate. Any capacity below full capacity results in lower survival rates due to ventilator shortages. The allocation rate is calculated by dividing the total number of ventilators allocated by the total number of the ventilators requested. The bar plot associated with each panel indicates the area under the survival-capacity curve and allocation-capactiy curve, respectively, where a larger value indicates that the protocol can save more lives across different levels of shortages. Notably, the MP and SOFA curves exhibit overlap, indicating similar allocation patterns. Similarly, the lottery and youngest curves show close proximity, as do our two TxDDQN configurations.}
    \label{fig:ASCC}
\end{figure}

\begin{figure}[!t]
    \centering
    \includegraphics[width=\textwidth]{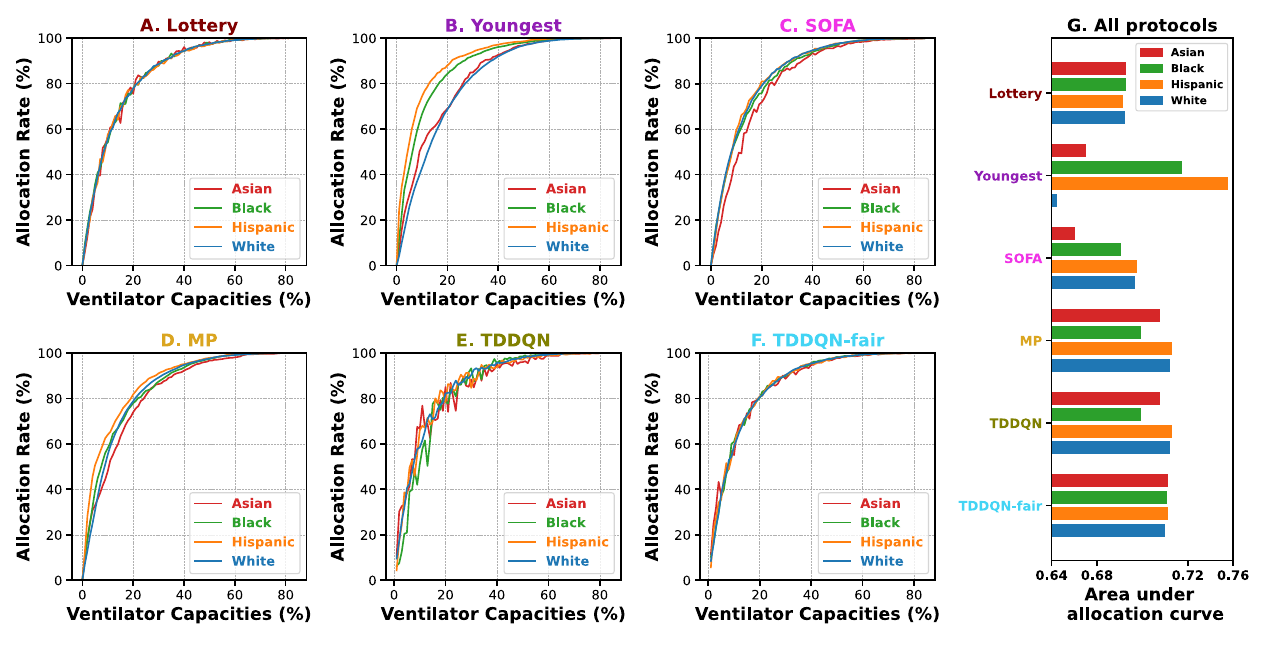}
    \caption{Allocation Rates across different protocols and ethnoracial groups when daily reassessment is considered. Panels A to F depict the ethnoracial differences in allocation rates when applying different protocols. Each panel focuses on a specific protocol and shows how the allocation rates vary across ethnoracial groups. Panel G presents the area under the allocation-capacity curve across all protocols and all ethnoracial groups.}
    \label{fig:AACC}
\end{figure}

Similarly to the settings without reassessment, we present our results in Table \ref{tab:stab2}, Figure \ref{fig:ASCC}, and Figure \ref{fig:AACC}. We find that our proposed TxDDQN-fair method improves the survival rate compared to all baseline protocols and achieves fair allocation across ethnoracial groups. The TxDDQN-fair configuration, with fairness rewards, yields higher survival rates than TxDDQN alone and results in a more equitable distribution (with the second-highest DPR, only smaller than the lottery protocol). Furthermore, when considering reassessment, the survival rates of all protocols are higher than when applied them in the non-reassessment setting. This suggests that withdrawing ventilators and reallocating them to patients with higher priority can save more lives. However, ethical considerations should be taken into account, and the decision of whether to include reassessment should be left to the physicians and stakeholders. 

\begin{table}[hbtp]
\centering
\caption{Impact of triage protocols on survival, fairness, and allocation rates with limited ventilators (B = 40) corresponding to about 50\% scarcity, when daily reassessment are considered. TxDDQN and TxDDQN-fair are without and with fairness rewards, respectively. Fairness metric is demographic parity ratio (DPR, the ratio between the smallest and the largest allocation rate across patient groups, 100\% indicating non-discriminative). Standard deviations are from 10 experiments with different seeds. We bold the ethnoracial group results with the highest survival rate, DPR, and overall allocation rates. We underline the metrics that fall within one standard deviation of the best result.}
\label{tab:stab2}
\resizebox{\columnwidth}{!}{%
\begin{tabular}{@{}l|l|l|lllll@{}}
\toprule
                  & \textbf{Performance} & \textbf{Fairness} & \multicolumn{5}{l}{\textbf{Allocation Rates}}                            \\ \cmidrule(l){2-8} 
                  & Survivals            & DPR, \%           & Overall, \%  & Asian, \%    & Black, \%    & Hispanic, \% & White, \%    \\ \midrule
\textbf{Lottery}    & 83.11 ± 0.40          & \textbf{98.95 ± 0.54} & 94.37 ± 0.11          & 94.74 ± 0.79 & 94.32 ± 0.34 & 94.20 ± 0.55 & 94.41 ± 0.12 \\
\textbf{Youngest} & 81.38 ± 0.10         & 94.63 ± 0.12      & 93.47 ± 0.10 & 92.72 ± 0.17 & 96.08 ± 0.05 & 97.18 ± 0.03 & 91.96 ± 0.02 \\
\textbf{SOFA}     & 87.88 ± 0.30         & 98.02 ± 0.63      & 94.38 ± 0.58 & 92.96 ± 0.60 & 93.83 ± 0.15 & 94.50 ± 0.35 & 94.77 ± 0.13 \\
\textbf{MP}       & 88.46 ± 0.13         & 96.51 ± 0.50      & 94.41 ± 0.29 & 92.05 ± 0.49 & 93.70 ± 0.20 & 95.37 ± 0.14 & 94.63 ± 0.05 \\ \midrule
\textbf{TxDDQN}    & \underline{90.44 ± 0.19}         & 95.08 ± 0.68      & \underline{95.18 ± 0.76} & 94.05 ± 1.22 & 94.66 ± 1.75 & 92.73 ± 1.35 & 96.02 ± 0.36 \\
\textbf{TxDDQN-fair} & \textbf{90.53 ± 0.17} & \underline{98.51 ± 0.61}          & \textbf{95.25 ± 0.72} & 94.42 ± 0.81 & 95.21 ± 0.57 & 94.87 ± 0.57 & 95.49 ± 0.81 \\ \bottomrule
\end{tabular}%
}
\end{table}
\newpage

\clearpage
\newpage

\subsection{A.6 No immediate death due to ventilation shortage}\label{sec:nodeath}
In the main text, we addressed a scenario in which patients die immediately if they request ventilators, but cannot be allocated due to resource shortages. This assumption was grounded in the understanding that ventilators are life-saving devices, and patients in critical care units are critically ill. Additionally, we are unable to obtain or simulate patient trajectories when a patient needs ventilators but is unable to receive one, as collecting this type of data is impractical. Thus, it is impossible to precisely estimate which patients or how many will die due to the lack of ventilation.

In this section, we make an effort to approximate the probability of mortality resulting from a lack of ventilation using a numerical value between 0 and 1. It's important to emphasize that these experiments should only be interpreted as a means to validate the generalizability of our proposed models. The exact probability can only be discerned by healthcare professionals and may vary depending on various factors such as healthcare resource availability, pandemic stages, seasonal variations, geographical locations, and other contextual considerations.

The results presented in Table \ref{tab:earlyD} clearly demonstrate that our proposed model surpasses all other protocols in terms of survival rates. It also achieves fair allocations that are on par with the lottery protocol and significantly outperforms the other three protocols. These superior outcomes hold true across various levels of the probability of death due to ventilation shortages, validating the versatility and applicability of our proposed model across various levels of urgency in healthcare resources allocation.

\begin{table}[h]
\centering
\caption{Impact of triage protocols on survival and fairness allocation rates with limited ventilators (B = 40) corresponding to about 50\% scarcity, where unmet ventilator requests do not result in immediate fatalities. The probability denotes an estimated rate of early patient deaths attributed to unmet ventilator needs. We scale the number of survivors to a range of {[}0, 100\%{]} to represent the survival rate. Fairness metric is demographic parity ratio (DPR, the ratio between the smallest and the largest allocation rate across patient groups, 100\% indicating non-discriminative). We bold the protocols with the highest survival rate and DPR at different levels of probability. We underline the metrics that fall within one standard deviation of the best result.}
\label{tab:earlyD}
\resizebox{\textwidth}{!}{%
\begin{tabular}{@{}c|cc|cc|cc|cc|cc@{}}
\toprule
 & \multicolumn{2}{l|}{} & \multicolumn{2}{l|}{} & \multicolumn{2}{l|}{} & \multicolumn{2}{l|}{} & \multicolumn{2}{l}{} \\
 & \multicolumn{2}{l|}{\multirow{-2}{*}{Lottery}} & \multicolumn{2}{l|}{\multirow{-2}{*}{Youngest}} & \multicolumn{2}{l|}{\multirow{-2}{*}{SOFA}} & \multicolumn{2}{l|}{\multirow{-2}{*}{MP}} & \multicolumn{2}{l}{\multirow{-2}{*}{TxDDQN-fair}} \\ \cmidrule(l){2-11} 
\multirow{-3}{*}{\begin{tabular}[c]{@{}c@{}}Probability of \\ Death due to \\ No Ventilation\end{tabular}} & Survival & DPR & Survival & DPR & Survival & DPR & Survival & DPR & Survival & DPR \\ \midrule
1 & 82.86 & \textbf{98.98} & 81.34 & 94.65 & 87.74 & 98.04 & 88.40 & 96.46 & \textbf{90.54} & \underline{98.54} \\
0.9 & 83.60 & \textbf{98.86} & 82.29 & 94.36 & 88.13 & 97.86 & 88.83 & 96.43 & \textbf{90.80} & \underline{98.42} \\
0.8 & 84.14 & \textbf{98.77} & 83.10 & 93.99 & 88.66 & 97.73 & 89.35 & 96.42 & \textbf{91.47} & \underline{98.40} \\
0.7 & 84.87 & \textbf{98.72} & 84.25 & 93.52 & 89.11 & 97.66 & 89.82 & 96.19 & \textbf{91.83} & \underline{98.36} \\
0.6 & 85.38 & \textbf{98.63} & 85.19 & 93.02 & 89.73 & 97.63 & 90.27 & 96.06 & \textbf{92.04} & \underline{98.24} \\
0.5 & 86.37 & \textbf{98.52} & 86.35 & 92.55 & 90.50 & 97.58 & 90.87 & 95.95 & \textbf{92.44} & \underline{98.23} \\
0.4 & 87.65 & \textbf{98.47} & 87.99 & 91.73 & 91.31 & 96.74 & 91.80 & 95.22 & \textbf{93.72} & \underline{98.23} \\
0.3 & 89.13 & \underline{98.04} & 89.76 & 90.46 & 92.32 & 96.46 & 92.77 & 94.86 & \textbf{94.15} & \textbf{98.18} \\
0.2 & 91.02 & \underline{97.87} & 91.69 & 88.93 & 93.84 & 95.60 & 94.18 & 94.21 & \textbf{95.42} & \textbf{98.14} \\
0.1 & 94.27 & \underline{97.49} & 94.80 & 87.19 & 95.91 & 94.12 & \underline{96.12} & 94.00 & \textbf{96.36} & \textbf{98.10} \\ \bottomrule
\end{tabular}%
}
\end{table}

\newpage

\subsection{A.7 Limitations}\label{sec:limitation}
In this study, we acknowledge several limitations that can serve as valuable directions for advancing our research. Firstly, our model was developed using data solely from one health system. Although we carefully divided the data into training, testing, and validation sets, utilizing both simulator data and real-world data, the generalizability of our findings may be limited. To address this, it would be beneficial to obtain data from multiple sites, allowing us to assess and discuss the model's performance across different healthcare settings. Secondly, our current model requires separate training for each specific ventilator shortage level. We aim to overcome this limitation by developing a more generalized model that can adapt with minimal changes as the capacity fluctuates. This would enhance the model's flexibility and practicality in real-world scenarios. Thirdly, our model has the potential to extend its application to other critical healthcare resources such as ECMO (Extracorporeal Membrane Oxygenation) and ICU beds. However, due to data limitations, we are constrained to conducting experiments and evaluations on these resources only when data becomes available. Expanding our research to include these resources would provide a more comprehensive understanding of their utilization and optimization. Fourthly, we introduced a Q network parametrization to address the challenging complexity inherent in the classic Q network. The effectiveness of this parametrization is underscored by our robust empirical results. However, establishing an approximation bound for this inherently complex combinatorial optimization problem poses a highly nontrivial but intriguing challenge. We plan to explore this aspect in our future theoretical studies. By addressing these limitations and exploring the proposed directions, we can enhance the robustness, applicability, and effectiveness of our study.

%% file: tables/table3.tex
\begin{table}[ht]
\centering
\caption{ Summary statistics of our dataset. For rows \texttt{Train}, \texttt{Validation}, \texttt{Test} and \texttt{Overall}, $N(P\%)$ are the number of patients and the percentage out of all races. For rows \texttt{Female} and \texttt{Deaths}, $N(P\%)$ are the number of patients and the percentage of females and deaths within that race. $\pm$ denotes std. \texttt{Vent days} is the number of days a patient has been allocated a ventilator. }
\vskip 0.15in
\resizebox{0.85\columnwidth}{!}{%
\begin{tabular}{l|llll|l}
\toprule
      & Asian       & Black        & Hispanic      & White      & All races  \\
      \midrule
Train & 206 (3.8\%) & 871 (16.0\%) & 668 (12.2\%)  & 3381 (62.0\%) &  5455\\
Validation   & 34 (3.2\%)  & 161 (15.4\%) & 109 (10.4\%)  & 663 (63.3\%)  &  1047\\
Test  & 239 (4.5\%) & 704 (13.4\%) & 497 (9.4\%) & 3465 (65.7\%)  &  5271 \\
\midrule
Overall & 479 (4.1\%)  & 1736 (14.7\%) & 1274 (10.8\%) & 7509 (63.8\%)  & 11773 \\
\midrule
Age & 62.3 ± 16.0 & 57.5 ± 16.2 & 55.9 ± 15.8 & 64.2 ± 14.7 & 62.0 ± 15.5 \\
Female & 184 (38.4\%) & 814 (46.9\%) & 495 (38.9\%) & 2848 (37.9\%) & 4630 (39.3\%) \\
Deaths & 122 (25.5\%) & 409 (23.6\%) & 358 (28.1\%) & 1870 (24.9\%) &  2980 (25.3\%) \\
Vent days & 4.9 ± 6.1 & 5.9 ± 6.7 & 6.5 ± 7.3 & 4.3 ± 5.3 &  4.8 ± 5.9 \\
\bottomrule
\end{tabular}
}
\label{tab:dataset}
\end{table}